\documentclass[conference]{IEEEtran}
\usepackage{graphicx}
\usepackage{subfigmat}% matrices of similar subfigures, aka small mulitples

\begin{document}

\title{Motion Primitives for Robotic Flight Control}

\author{\authorblockN{Bar{\i}\d{s} E. PERK}
\authorblockA{Nonlinear Systems Laboratory\\Massachusetts Institute of Technology\\
Cambridge, Massachusetts, 02139, USA\\
Email: bperk@mit.edu} \and
\authorblockN{J.J.E. Slotine}
\authorblockA{Nonlinear Systems Laboratory\\Massachusetts Institute of Technology\\
Cambridge, Massachusetts, 02139, USA\\
Email: jjs@mit.edu} }

\maketitle

\begin{abstract}

We introduce a simple framework for learning aggressive maneuvers in
flight control of UAVs. Having inspired from biological environment,
dynamic movement primitives are analyzed and extended using
nonlinear contraction theory. Accordingly, primitives of an observed
movement are stably combined and concatenated. We demonstrate our
results experimentally on the Quanser Helicopter, in which we first
imitate aggressive maneuvers  and then use them as primitives to
achieve new maneuvers that can fly over an obstacle.

\end{abstract}

\section{Introduction}

  The role of UAVs (Unmanned Aerial Vehicles) has gained significant
  importance in the last decades.  They have many advantages (agility,
  low surface area, ability to work in constrained or dangerous
  places) over their conventional precedents. In addition, current UAVs
  are more biologically-inspired in terms of shape and performance
  because of the improvements in electronics and propulsion. Unfortunately,
  we are still far away from using their capacity at the fullest. This  is mostly related
  with the weakness of current control algorithms against
  high-dimensional and   nonlinear environments. In this sense, generating aggressive maneuvers is interesting and hard
  to accomplish.

In this paper, our approach to solve this issue is designed in view
of the experiments on frogs and monkeys which suggest that we are
faced with an inverse-kinematics algorithm that adapts to the
environment and changes in a sequence of target points irrespective
of the initial conditions. In theory, we analyzed dynamic movement
primitives (DMPs)\cite{schaal2002} and combined them using
contraction theory. In experiments, obstacle avoidance DMP of a
human-piloted flight data is segmented into parts and combined at
different initial points to achieve maneuvers against different
obstacles on different locations. Background of our work is briefly
detailed below.

\subsection{Background}
\subsubsection{Imitation Learning}

"By three methods we may learn wisdom: first, by reflection, which
is noblest; second, by imitation, which is easiest; and third, by
experience, which is the most bitter." (Confucius) Imitation takes
place when an agent learns a behavior by observing the execution of
that behavior from a teacher~\cite{kuniyoshi}. Imitation is not
inherent to humans. It is also observed in animals. For example,
experiments show that kittens exposed to adult cats manipulate
levers to retrieve food much faster than the control
group~\cite{galef}.

There has been a number of applications on imitation learning in the
field of robotics. Studies on locomotion \cite{biped1,
biped2,billard}, humanoid robots \cite{juggling,
movement},\cite{mataric}, \cite{kuniyoshi2}, and human-robot
interactions \cite{mussa2000b, nicolescu} have used imitation
learning or movement primitives.  The emphasis on these studies is
on primitive derivation and movement classification \cite{mataric1};
combinations of the primitives \cite{nakaoka, slotinecdc, Burridge,
Inamura,khatib,giese1} and primitive models \cite{schaal,
mussa_neural, mussa_nonlinear,giese1} in order to extract behaviors.

\subsubsection{Aggressive Maneuvers}

Aggressive control of autonomous helicopters represents a
challenging problem for engineers. The challenge owes itself to the
highly nonlinear and unstable nature of the dynamics along with the
nonlinear relations for actuator saturation. Nevertheless, we can
find successful unmanned helicopter examples
\cite{feron,frazzoli1,andrew,bagnell,vladislav,mahony,shakernia,shim,mettler}
in the literature. However, model helicopters controlled by humans
can achieve considerably more complex and aggressive maneuvers
compared to that can be done autonomously with the state of the art.
In~\cite{frazzoli2}, it is observed that after several repetitions
of the same maneuver, performed by a human, generated trajectories
are similar and the control inputs are well-structured and
repetitive. Hence, it is intuitive to focus on understanding human's
maneuvers to find proper algorithms for unmanned control.

\subsubsection{Biological Motivation}

In their experiment with deafferented and intact monkeys, Bizzi
\cite{bizzi1979} found that a certain movement can be executed
regardless of initial conditions, emphasizing the importance of
feedback control. In particular, they have shown that the control
variable is the equilibrium state of the agonist and antagonist
muscles. Same experimental setup is again used to characterize the
trajectory of the motion in \cite{bizzi1986}.  Their results
additionally suggest that movement called "virtual trajectory" is
composed of more than one equilibrium point and central nervous
system uses the stability of the lower level of the motor system to
simplify the generation of movement primitives\cite{bizzi1986}.

 Bizzi \cite{bizzi 1991} and Mussa-Ivaldi
\cite{mussa1}'s experiments on frogs provide us with further clues
in understanding movement primitives. They microstimulated spinal
cord and measured the forces at the ankle.  Having repeated this
process with ankle replaced at nine to 16 locations, they observed
that collection of measured forces always converges to a single
equilibrium point. In their model, inverse kinematics plays a
crucial role in achieving the endpoint trajectory (see Mussa-Ivaldi
\cite{mussa2000}).

\section{Analysis of DMP}

This section outlines the analysis of the DMP algorithm using
contraction theory.

\subsection{DMP Algorithm}

   DMP is a trajectory generation algorithm which interpolates between the start
    and end points of a path based on learning.  The system can be
    represented by

\begin{equation}
\tau\dot{z}=\alpha_z(\beta_z(g-y)-z) \label{eq:1}
\end{equation}
\begin{equation}
\tau\dot{y}=z+f \label{eq:2},
\end{equation}
where $y$, $\dot{y}$ and $\ddot{y}$  characterize the desired
trajectory, $\alpha_z$ and $\beta_z$ are time constants, $\tau$ is a
temporal scaling factor, $g$ is the desired end point. In addition,
the canonical system is given by

\begin{equation}
\tau\dot{v}=\alpha(\beta_z(g-x)-v) \label{eq:4}
\end{equation}
\begin{equation}
\tau\dot{x}=v\label{eq:5},
\end{equation}
In general, assuming that the $f$-function is zero, system will
converge to $g$ exponentially.  The goal of the DMP algorithm is to
modify this exponential path so that the $f$-function makes the
system non-linear and allows us to generate desired trajectories
between the origin and the $g$ point.

The $f$-function is a normalized linear combination of Gaussians
which helps to approximate the final trajectory, i.e. it has the
general form
\begin{equation}
f(x,v,g)=\frac{\sum_{i=1}^N\Psi_iw_iv}{\sum_{i=1}^N\Psi_i},
\label{fDef}
\end{equation}
where \begin{equation} \Psi_i=exp\{-h_i(x/g-c_i)^2\}.
\end{equation}

\subsection{Rhythmic DMPs}

The DMP algorithm can also be extended to the rhythmic movements
\cite{ijspeert_rhythmic} by changing the canonical system with the
following:

\begin{eqnarray}
\tau\dot{\phi}&=&1 \label{rcp1}\\
\tau\dot{r}&=&-\mu(r-r_0) \label{rcp2} \hspace{1.5 cm} \mu>0
\end{eqnarray}
where $\phi$ corresponds to $x$ in Eq. \ref{eq:4} as a temporal
variable. Similar to the discrete system, control policy: \\
\begin{eqnarray}
\tau\dot{z}&=&\alpha_{z}(\beta_z(y_m-y)-z) \\
\tau\dot{y}&=&z+f\\
f(x,v,g)&=&\frac{\sum_{i=1}^N\Psi_iw_{i}^T\tilde{v}}{\sum_{i=1}^N\Psi_i} \label{f_function}\\
\psi&=&exp\{h_i(\cos(\phi-c_i)-1)\}
\end{eqnarray}
where $y_m$ is a basis point for learning and
$\tilde{v}=[x=r\cos(\phi),y=r\sin(\phi)]^T$.

\subsection{Learning of primitives using DMPs}
 Learning aspect of the algorithm comes into play with the
computation of the weights ($w_i$) of the Gaussians. Weights are
derived from Eq.\ref{eq:1} and Eq.\ref{eq:2} using the training
trajectory $y_{demo}$ and $\dot{y}_{demo}$ as variables . Once the
parameters of the $f$-function are learned, then DMP can simply be
used to generate the original trajectory. As detailed below, spatial
and temporal shifts are achieved by adjusting the $g$ and $\tau$
respectively.

\begin{itemize}

\item \emph{Spatial adjustments:}
The first system [Eq.(\ref{eq:1}), Eq.(\ref{eq:2})]  can be seen as
a linear system. It is due to the fact that variable $v$ in
$f$-function is only multiplied  by time-varying constant. Hence, we
can say that output ($y$) is simply scaled by $g$ from
superposition.

\item \emph{Temporal adjustments:} The second system [(Eq.(\ref{eq:4}) Eq.(\ref{eq:5})] is simply
linear. In addition, f function is linear because the multiplier
$\Psi_i$ is a time-varying constant, temporally scaled by $\tau$.
Thus, from linearity, we can say that temporal adjustments of the
whole system is carried out by just changing the variable $\tau$.
 %(See Fig. \ref{learning1}).

\end{itemize}
These arguments can also be extended to the rhythmic DMPs for
modulations.
%\begin{figure}
%\begin{minipage}[t]{4.3cm}
%\includegraphics[width=0.9\textwidth]{illustr/figure_x} \caption{variable x with $g$=3,6 }
% \label{learning1}
%\end{minipage}
%\hfill
%\begin{minipage}[t]{4.3cm}
%\includegraphics[width=0.9\textwidth]{illustr/figure_y}
%\caption{variable y with $g$=3,6 } \label{learning2}
%\end{minipage}
%\hfill
%\end{figure}

\subsection{Analysis of DMP Using Contraction Theory}

The basic theorem of contraction analysis~\cite{slotine3} is stated
as

\noindent \textbf{Theorem (Contraction)}\emph{Consider the
deterministic system}

\begin{eqnarray}
\dot{x}=f(x,t)
\end{eqnarray}

\noindent\emph{where $\textbf{f}$ is a smooth nonlinear function. If
there exist a uniformly invertible matrix %$\textbf{\Theta(x,t)}$ that the
associated generalized Jacobian matrix}

\begin{eqnarray}
F=(\dot{\Theta}+\Theta\frac{\partial f}{\partial x})\Theta^{-1}
\end{eqnarray}

\noindent \emph{is uniformly negative definite, then all system
trajectories converge exponentially to a single trajectory, with
convergence rate $|\lambda_{max}|$, where $\lambda_{max}$is the
largest eigenvalue of the symmetric part of F. The system is said to
be contracting.} \\

 Basically, a nonlinear time-varying dynamic system is called
contracting if initial conditions or temporary disturbances are
forgotten exponentially fast, i.e., if trajectories of the perturbed
system return to their nominal behavior with an exponential
convergence rate.  It turns out that relatively simple conditions
can be given for this stability-like property to be verified.
Furthermore this property is preserved through basic system
combinations, such as parallel combinations, feedback combinations,
and series or hierarchies, yielding simple tools for modular design.
For linear time-invariant systems, contraction is equivalent to
strict stability.

Consider a system

\begin{eqnarray}
 \frac{d}{dt}\left[
\begin{array}{c}
\delta z_1 \\
\delta z_2
\end{array}
\right] & = & \left[
\begin{array} {cc}
F_{11} & 0\\
F_{21} & F_{22}
\end{array}
\right]
\left[
\begin{array} {c}
\delta z_1\\
\delta z_2
\end{array}
\right] \label{equation13}
\end{eqnarray}\\

\noindent where $z_1$ and $z_2$ represent the first and  the second
system  of DMP and the $\delta z_i$ represent associated
differential displacements (see \cite{slotine3}) . Equation
(\ref{equation13}) display a hierarchy of contracting systems, and
 furthermore since $F_{21}$ is bounded by construction of $f$, the
 whole system globally exponentially converges to a single trajectory \cite{slotine3}.

%Hierarchical contraction can also be extended to rhythmic DMPs, since
%the derivatives of  $x=r\cos(\phi)$ and $y=r\sin(\phi)$ are
%contracting.
We can also extend the hierarchical contraction property to the
rhythmic DMPs, since the canonical system, which is shown below is
contracting.
\begin{eqnarray}
\tau\dot{x}=  -\mu (x-x_0) - y\hspace{1 cm} \mu>0\\
\tau\dot{y}=  -\mu (y-y_0) + x \hspace{1.9cm}
\end{eqnarray}

Although the system will eventually contract to the $g$ point, there
will be a time delay due to the hierarchy between second and the
first system. We can decrease this delay by increasing the number of
weights in our equation.

Using contraction theory, stability of the DMPs can  be analyzed.
Once the original trajectory is mapped into the DMP, the system
behaves linearly for a given input-output relation as shown before.
Moreover, contraction property guarantees the convergence into a
single trajectory. From linearity, it is easy to show that learning
the trajectories is not constrained by the stationary goal points
that do not have a velocity components, which are required for
equilibrium points in virtual trajectories.

\section{Coupling of DMPs Using Contraction Theory}

In this section, we use partial contraction theory \cite{slotine4}
to couple DMPs. One-way coupling configuration of contraction theory
allows a system to converge to its coupled pair smoothly. Theory for
the one-way coupling states the following two systems:

\begin{equation}
\dot{x}_1=f(x_1,t)
\end{equation}
\begin{equation}
\dot{x}_2=f(x_2,t)+u(x_1)-u(x_2)
\end{equation}

In a given formula, if $f-u$ is contracting, then $x_2\rightarrow
x_1$ from any initial condition.

A typical example for one way coupling is an observer design while
 the first system represents the real plant and the second system
represents the mathematical model of the first system. The states of
the second system will converge to the states of the first system
and result in the robust estimation of the real system states.
However, for our experiments, we interpret contraction as to imitate
the transition between two states. It will be shown in section IV
how the end of one trajectory becomes the initial condition of the
second trajectory and contraction accomplishes the smooth
transition.

In DMPs, we couple the two systems using the following equations:
\begin{equation}
\ddot{y}_1=g_1-y_1-\dot{y_1}+f(y_1) \label{sistem1}
\end{equation}
\begin{equation}
\ddot{y}_2=g_2-y_2-\dot{y_2}+u(y_1)-u(y_2) \label{sistem2}
\end{equation}
\begin{equation}
u(x_i)=g_i+f(x_i)\end{equation}
\begin{equation}
f(x,v,g)=\frac{\sum_{i=1}^N\Psi_iw_iv}{\sum_{i=1}^N\Psi_i}
\end{equation}
\begin{equation}
\Psi_i=exp\{-h_i(x/g-c_i)^2\}
\end{equation}

\begin{figure}
\begin{center}
\includegraphics[width=3.5 in]{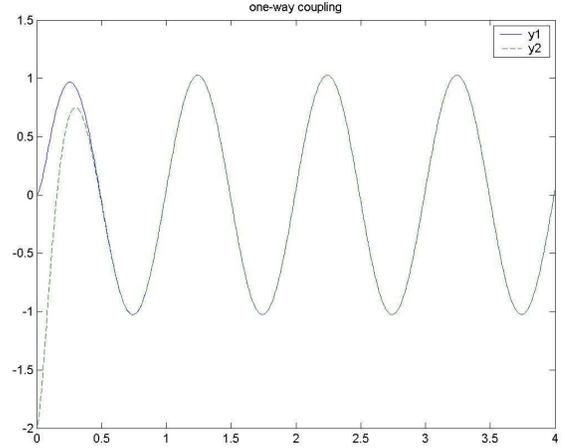}
\caption{One-way coupling of a rhythmic DMPs}\label{one2}
\end{center}
\end{figure}

  A toy example of the equations listed above can be
  seen in Fig. \ref{one2}. In this setting, $y_2$  is the first
  trajectory primitive, which contracts to $y_1$  -- the second trajectory primitive.

   One-way coupling has many advantages as a method over its
    precedents:

     In \cite{giese1}, trajectories are achieved by
    simply
    stretching the original trajectory in its coordinates and
    there is a direct relation between initial and end points. Also,
    there are discontinuities in terms of derivatives
    of the trajectory at the transition regions between primitives. Giese
    \cite{giese2}
    solves the problem of discontinuities by first
    taking the derivatives of the original trajectories, then
    combining     the derivatives, and finally
    integrating them again using initial conditions.
    However, this method adversely affects the accuracy of the
     trajectories. Hence, our method improves on \cite{giese1} and
     \cite{giese2} by generating more accurate trajectories
     independent of initial points.

In \cite{andrew2}, snapshots of the pilot's maneuvers are taken and
evaluated as noisy measurements of hidden and true trajectory. In
their model, time indexes are used for the comparison of expert's
demonstrations. Maximization of the joint likelihood of
demonstrations are achieved through trajectory learning algorithms.
As was done in \cite{andrew}, Locally Weighted Learning is used for
learning system dynamics close to trajectories. Moreover, desired
trajectories are supervised by adding information specific to each
maneuver. With the help of feasible trajectory, optimal controller
and system dynamics along the maneuver, they achieved remarkable
results on model helicopters.  However, finding hidden trajectory
requires noteworthy computational performance where they smooth out
data to emphasize the similarities. In addition, their algorithm
applies only for mimicking demonstrations. In our algorithm,
learning the hidden and true trajectory of maneuvers can simply be
done by comparing the weights of DMPs (see \cite{schaal2002}). It is
also easier to manipulate DMPs by changing parameters ($\tau$ and
$g$) for new challenges.   Moreover, our method lies on the
background of biological experiments in such a way that it is
adaptable for further research.

      In general, we summarize the advantages for
      using dynamical systems as control policies as follows:

      \begin{itemize}\item It is easy to incorporate
    perturbations to dynamical systems. \item  It is easy to represent the primitives.
    \item
     Convergence
    to the goal position is guaranteed due to the attractor dynamics of DMP.
    \item It is
    easy to
    modify for different tasks.
    \item At the
    transition regions, discontinuities are avoided.
    \item Partial contraction theory forces the coupling
    from any initial condition.
    \end{itemize} Also in \cite{schaal2002}, Schaal's suggested system is driven between stationary
points.
    However, biological experiments suggest that we are faced with a "virtual
    trajectory"  composed of equilibrium points that has velocity
    components. For this reason, we showed that we can achieve this
    property by combining nonconstant points.

\section{Experiments on Helicopter}

Here, we apply the motion primitives on the helicopter.

\subsection{Experimental Setup}
\begin{figure}[h]
\centering
\includegraphics[width=2in]{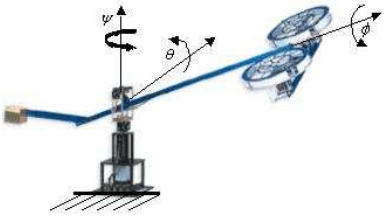}
\caption{Transverse momentum
distributions.\cite{masha}}\label{fig:quanser}
\end{figure}

We used Quanser Helicopter (see Figure \ref{fig:quanser}) in our
experiments. The helicopter is an under-actuated system having two
propellers at the end of the arm. Two DC motors are mounted below
the propellers to create the forces which drive propellers. The
motors' axes are parallel and their thrust is vertical to the
propellers. We have three degrees of freedom (DOF): pitch (vertical
movement of the propellers), roll (circular movement around the axis
of the propellers) and travel (movement around the vertical base) in
contrast with conventional helicopters with six degrees of freedom.

\begin{figure}
\begin{minipage}[t]{5cm}
\includegraphics[width=0.9\textwidth]{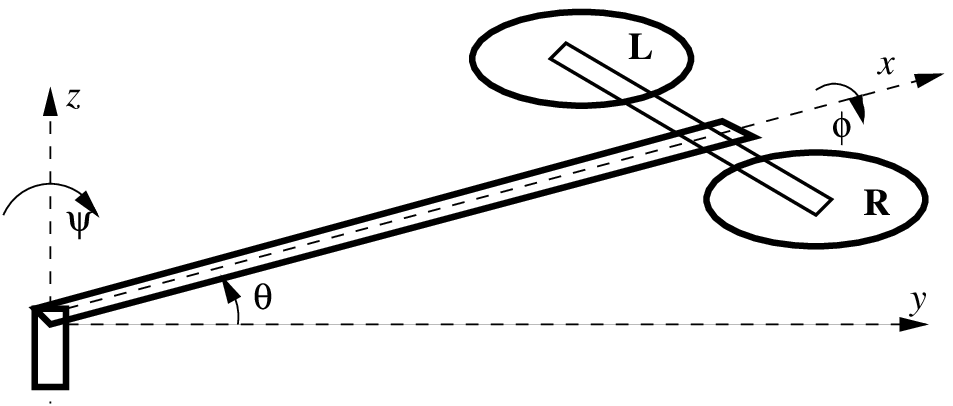} \caption{Schematic
of the 3DOF helicopter.\cite{masha}} \label{fig:heli_draw}
\end{minipage}
\hfill
\begin{minipage}[t]{3.5cm}
\includegraphics[width=0.9\textwidth]{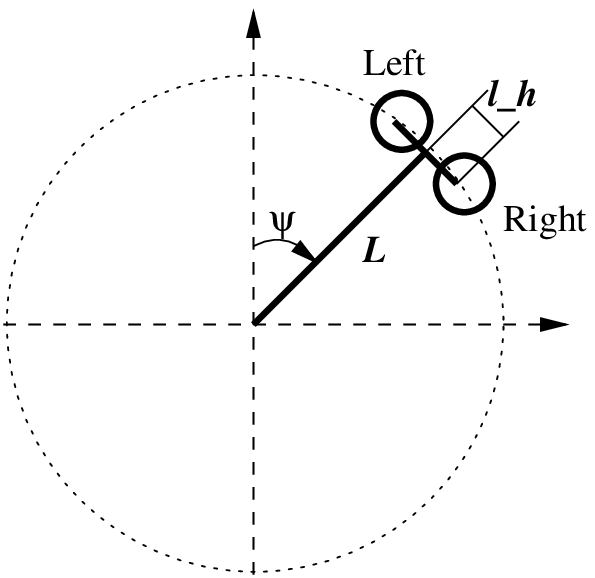}
\caption{Top view. \cite{masha}} \label{fig:topview}
\end{minipage}
\hfill
\end{figure}

  In system model\cite{masha}, the origin of our coordinate system is at the
  bearing and slip-ring assembly. The combinations of  actuators
  form the collective $(T_{col}=T_L + T_R)$ and cyclic
  $(T_{cyc}=T_L-T_R)$ forces which are used as inputs in our controller. %The pitch
%  and roll motions are controlled by collective and cyclic thrusts
%  respectively. Motion in travel angle is controlled by the
%  components of thrust. Positive roll results in positive change of
%  angle. %
The schematics of helicopter are shown in Figures
~\ref{fig:heli_draw}~and~\ref{fig:topview}.

  Let $J_{xx}$, $J_{yy}$, and $J_{zz}$ denote the
  moment of inertia of our system dynamics.
  For simplicity, we ignore the products of inertia terms.
 The equations of motion are as follows (cf. Ishutkina \cite{masha}):
\begin{displaymath}
J_{zz}\ddot{\psi}=(T_L+T_R)L\cos(\theta)\sin(\phi) \nonumber,
\end{displaymath}
\begin{displaymath}
-(T_L-T_R)l_h\sin(\theta)\sin(\phi)- \mbox{Drag}\label{Eq:psi}
\nonumber,
\end{displaymath}
\begin{displaymath}
J_{yy}\ddot{\theta}=-Mgl_{\theta}\sin(\theta+\theta_0) +
(T_L+T_R)L\cos(\phi)\label{Eq:theta} \nonumber,
\end{displaymath}
\begin{displaymath}
 J_{xx}\ddot{\phi}= - mgl_{\phi}\sin(\phi)+(T_L-T_R)l_h
\nonumber \label{Eq:phi} \label{eq:masha},
\end{displaymath}
where
\begin{itemize}
\item $M$ is the total mass of the helicopter assembly,
\item $m$ is the mass of the rotor assembly,
\item $L$ is the length of the main beam from the slip-ring pivot to
the rotor assembly,
\item $\psi$, $\theta$, $\phi$ are travel, pitch and roll angles
respectively.
\item $l_h$ is the distance from the rotor pivot to each of the
propellers,
\item
$\mbox{Drag}=\frac{1}{2}\rho(\dot{\psi}L)^2(S_0+S_{0}'\sin(\phi))L$,
\item $S_0$ and $S_{0}'$ are the effective drag coefficients times the
reference area and $\rho$ is the density of air.
\end{itemize}
It can be seen that the above system is nonlinear in the states, but
linear in terms of control inputs. In practice, we used feedback
linearization with bounded internal dynamics (see Bayraktar
\cite{selcuk2}) for a 3DOF helicopter, which tracks trajectories in
elevation and travel.

\subsection{Simulation \& Experimental Results}

In this section, we first describe our numerical simulation of
 the proposed primitive framework.  Second, we describe our actual
 experiment on the Quanser Helicopter.

\subsubsection{Trajectory Generation}

In experimental setup, we used an operator with a joystick
  to create aggressive trajectories to pass an obstacle. However, generating
aggressive
  trajectories with the joystick is a difficult task even for the
  operator.  Therefore, we designed an augmented control for the joystick to
  enhance the performance of the helicopter. In detail, we
  used "up" and "down" movements of the joystick to increase or decrease
  the $T_{col}$ that is applied to the actuators. For the "right" and
  "left" movements of the joystick, we preferred to control the roll angle
  using PD control.

\begin{figure}[hbt]
  \begin{center}
  \includegraphics[width=3.5in]{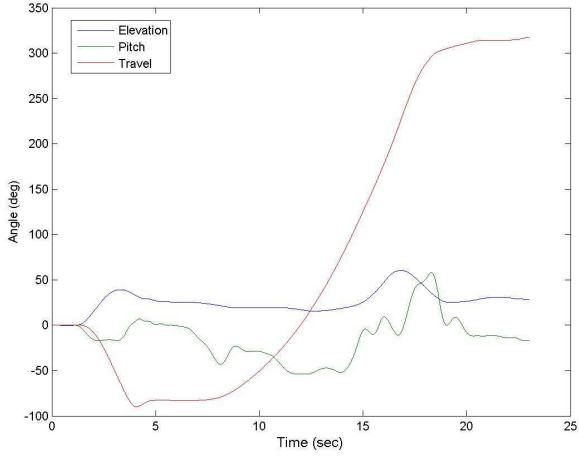}
  \end{center}
  \caption{Original maneuver achieved by an operator}
  \label{original}
\end{figure}

  In the original maneuver, the obstacle's distance and the highest
  point are in the coordinates where $\psi$ and $\theta$ angles are
  $220$ and $60$ respectively and the helicopter stops at the
  coordinates where $\psi=28$, $\theta=317$ and $\phi=-17$ (see Figure
  \ref{original}).

\subsubsection{Trajectory Learning}

\begin{figure}[hbt]
  \begin{center}
  \includegraphics[width=3.5in]{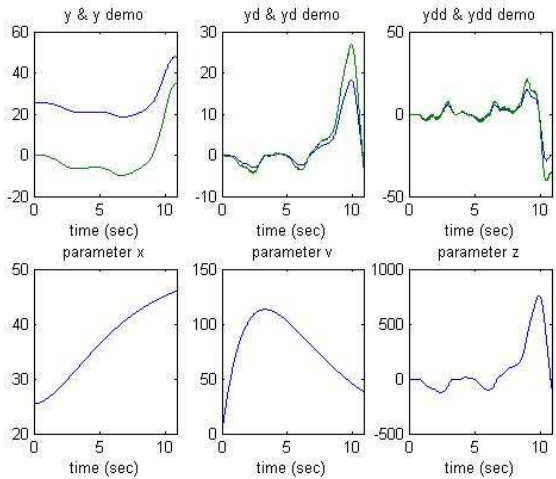}
  \end{center}
  \caption{Trajectory generation for the first primitive - pitch}
  \label{pitch1}
\end{figure}

\begin{figure}[hbt]
  \begin{center}
  \includegraphics[width=3.5in]{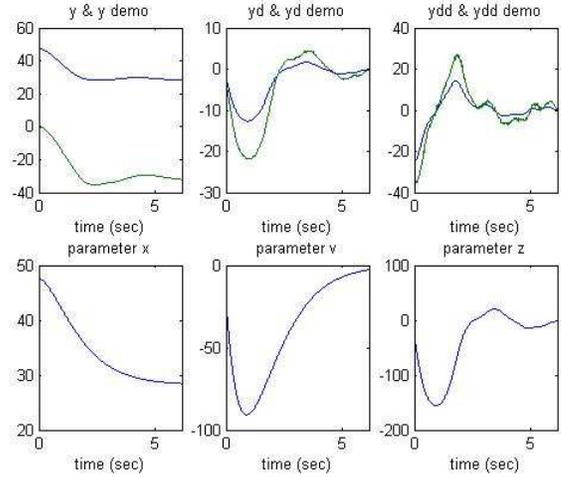}
  \end{center}
  \caption{Trajectory generation for the second primitive - pitch}
  \label{pitch2}
\end{figure}

From several demonstrations, it is observed that our operator
follows two distinct pattern to carry out the maneuver. Accordingly,
these two patterns suggest an equilibrium point at the top of the
obstacle. Therefore, to fly over different obstacles, the acquired
primitive is segmented into two primitives at the highest pitch
angle. Fig. \ref{pitch1} and Fig. \ref{pitch2} show the results of
DMP algorithm for the pitch angle.  The top left graphs are results
for pitch angles, where green lines represent the operator input for
the trajectories and blue lines represent the fittings that the DMP
computes for different start and end points. Hence, desired
trajectories in these graphs are not on top of the trajectories
generated by the operator. Other graphs show the time evolution of
the DMP parameters.


\subsubsection{Synchronization of primitives}

\begin{figure}[hbt]
  \begin{center}
  \includegraphics[width=4in]{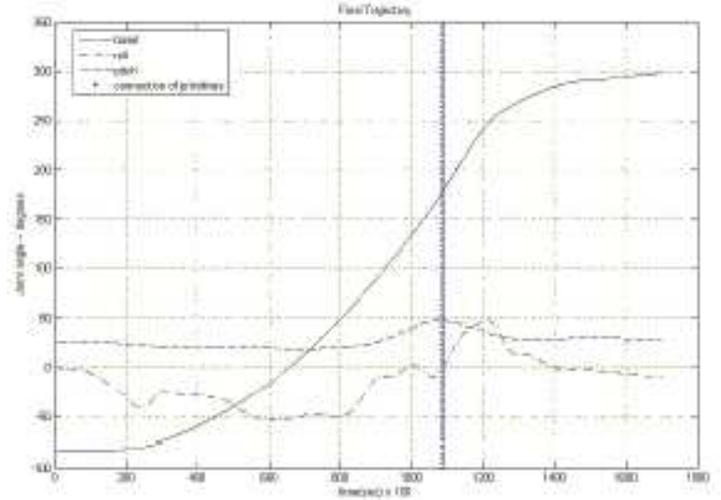}
  \end{center}
  \caption{Time evolution of primitive-1 and primitive-2 merged.}
  \label{fig:combined}
\end{figure}

The two primitives created in the previous sections are defined as
trajectories between certain start and end points. However, the end
point of the first trajectory does not necessarily matches with the
starting point of the second trajectory.  We use partial contraction
theory \cite{slotine4} to force the first trajectory to converge to
the second one. However, since we want to use the contraction as a
transition between two trajectories, coupling is enabled towards the
end of  first primitive.  Figure \ref{fig:combined} shows how the
two trajectories evolve in time. In the first primitive, the goal
positions of $\psi$ and $\theta$ angles are changed to $150^\circ$
and $50^\circ$ respectively, where original angles are
$\psi=220^\circ$ and $\theta=60^\circ$. In the second primitive, the
goal position of the $\psi$ angle is changed from $317^\circ$ to
$300^\circ$.

\subsubsection{Experiments on the Helicopter}

\begin{figure}[hbt]
  \begin{center}
  \includegraphics[width=3.5in]{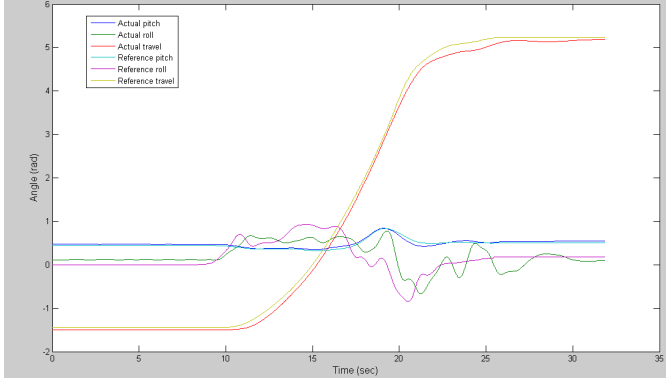}
  \end{center}
  \caption{Tracking performance of the helicopter.}
  \label{tracking}
\end{figure}

\begin{figure}[hbt]
  \begin{center}
  \includegraphics[width=3.5in]{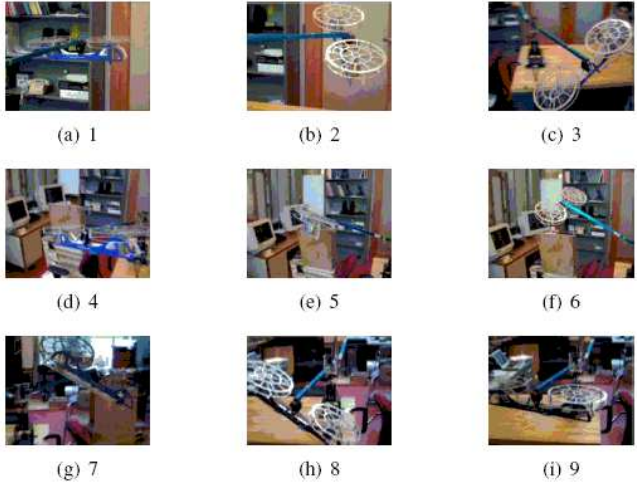}
  \end{center}
  \caption{Snapshots of the obstacle avoidance maneuver.}
  \label{fig:snapshots}
\end{figure}

 Tracking performance of the helicopter is shown in Figure \ref{tracking}. It is seen that
 the helicopter followed the desired
($\psi$ and $\theta$) angles almost perfectly. However, the
trajectory of the roll angle is a bit
 different than the desired since we
 control two parameters ($\psi$ and $\theta$) and the goal positions
 of the DMPs are different. But
we should highlight the fact that two roll trajectories follow the
same pattern. In figure, the last part of the roll trajectory
manifests an oscillation which can be prevented by roll control,
since the other parameters are almost constant. The tracking
performance can further be improved by applying discrete nonlinear
observers to get better velocity and acceleration values. Figure
\ref{fig:snapshots} shows snapshots of the maneuver.

%\begin{figure}
% \begin{subfigmatrix}{3}% number of columns
%  \subfigure[1]{\includegraphics[width=0.8in]{illustr/q1}}
%  \subfigure[2]{\includegraphics[width=0.8in]{illustr/q3}}
%  \subfigure[3]{\includegraphics[width=0.8in]{illustr/q4}}
%  \subfigure[4]{\includegraphics[width=0.8in]{illustr/q5}}
%  \subfigure[5]{\includegraphics[width=0.8in]{illustr/q6}}
%  \subfigure[6]{\includegraphics[width=0.8in]{illustr/q7}}
%  \subfigure[7]{\includegraphics[width=0.8in]{illustr/q8}}
%  \subfigure[8]{\includegraphics[width=0.8in]{illustr/q9}}
%  \subfigure[9]{\includegraphics[width=0.8in]{illustr/q10}}
% \end{subfigmatrix}
% \caption{Snapshots of the obstacle avoidance maneuver.}
% \label{fig:snapshots}
%\end{figure}

\section{ Extensions of DMP}

\subsection{Dynamical System with First-Order Filters}

DMP algorithm can be improved by replacing the first system with the
equations shown below:
\begin{eqnarray}
\tau\dot{y} + a_1 y = x \label{f1} \\
\tau \dot{x} + a_2 x = g+f \label{f2}
\end{eqnarray}
which is equivalent of
\begin{eqnarray}
\tau^2 \ \ddot{y} + \tau \ (a_1 + a_2) \ \dot{y} + a_1 a_2 \ y \ = \ g + f
\end{eqnarray}
By introducing two first-order filters, we  guarantee the stability
of the system against time varying parameters like $\tau(t)$ or
$g(t)$ . Since the system is linear without the $f$-function
(Eq.\ref{f_function}), we achieve learning and modulation properties
of DMP using the $f$ in either Eq.(\ref{f1}) or Eq.(\ref{f2}). For
further applications, we will use this model to generate primitives
for time-varying goal points.

\subsection{Generating New Primitives}
Experiments on frog's spinal cord \cite{bizzi 1991, mussa1,
ivaldi_linear} suggest that movement primitives can be generated
from linear combinations of vectorial force fields which lead the
limb of a frog to the virtual equilibrium points. In
\cite{ivaldi_linear}, it is also pointed out that vectorial
summation of two force fields with different equilibrium points
generate a new force field whose equilibrium point is at
intermediate location of the original equilibrium points. In this
perspective, we will use two methods to generate new primitives.

\subsubsection{Two-way Synchronization of DMPs}

 Consider a system
\begin{eqnarray}
\ddot{y}_1=f(y_1,t)+K(u(y_2)-u(y_1))\;\;\;\; \;\;\;\;\;\;\;\;\\
\ddot{y}_2=f(y_2,t)+K(u(y_1)-u(y_2))\;\;K>0
\end{eqnarray}
Where $y_1$ and $y_2$ represent the first and the second primitive
respectively. From partial contraction theory, we  say that $y_1$
and $y_2$ converge together exponentially, if $f-2Ku$ is
contracting. Since DMPs are already contracting, we  achieve
synchronization using contracting inputs. In Fig.\ref{synchronized
RCP} (top), new primitive is a linear combination of sine and cosine
primitives. Also in the same figure, coupling forces accounts for
oscillations before synchronization happens.

\begin{figure}[hbt]
  \begin{center}
  \includegraphics[width=3.5in]{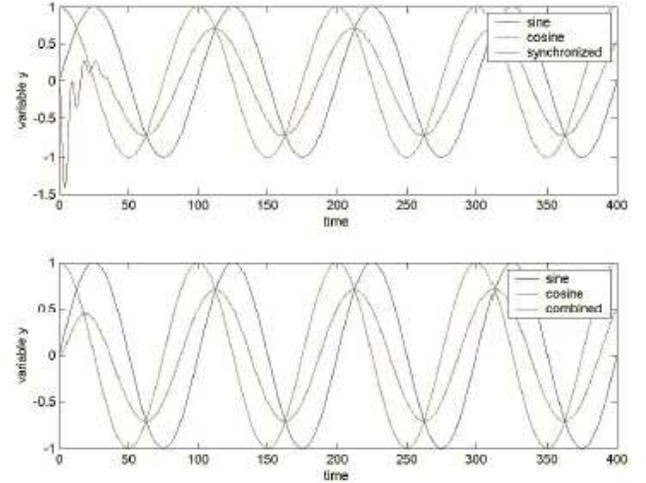}
  \end{center}
  \caption{\textbf{Top:}Synchronization of sine and cosine primitives. \textbf{Bottom:} New primitive
  generated by the linear combinations of weights from sine and cosine primitives}
  \label{synchronized RCP}
\end{figure}

\subsubsection{Combination of Primitives using Weights}

In DMPs, as it was shown before, system behaves linearly and
superposition applies. Therefore, in the  $f$-function , linear
combination of the weights from different primitives  produce linear
combination of primitives. For rhythmic DMPs, as an example, we
combine the weights of the sine and cosine primitives ($w_{new}=
0.5w_{sine}+0.5w_{cosine}$) to generate a new primitive (See Fig.
\ref{synchronized RCP} (bottom)). However for a regular DMP, we can
not achieve the desired trajectories although we have linearity
which is because input "$g$" point is not compatible with the
weights changing with respect to the couplings. For this reason, we
will simply modify the equations in our later research.

\section{Conclusion}

In this paper, we use a novel approach, inspired by biological
experiments and humanoid robotics, which uses control primitives to
imitate the data taken from human-performed obstacle avoidance
maneuver. In our model, DMP computes the trajectory dynamics so that
we can generate complex primitive trajectories for given different
start and end points, while one-way coupling ensures smooth
transitions between primitives at the equilibrium points. We
demonstrate our algorithm with an experiment.  We generate a
complex, aggressive maneuver, which our helicopter could follow
within a given error bound with a desired speed. Future research
will be conducted on different combinations of primitives using
partial contraction theory. We expect these techniques to be
particularly useful when the system dynamic models are very coarse,
as e.g. in the case of flapping wing systems and new bio-inspired
underwater vehicles.

\section*{Acknowledgment}

We extend our warm thanks to Prof. E. Feron and his PhD. student S.
Bayraktar for the opportunity to use their Quanser helicopter.

\end{document}